\documentclass[letterpaper]{article} 
\usepackage[draft]{aaai2026}  
\usepackage{times}  
\usepackage{helvet}  
\usepackage{courier}  
\usepackage[hyphens]{url}  
\usepackage{graphicx} 
\usepackage{natbib}  
\usepackage{caption} 
\usepackage{algorithm}
\usepackage{algorithmic}

\usepackage{newfloat}
\usepackage{listings}
\DeclareCaptionStyle{ruled}{labelfont=normalfont,labelsep=colon,strut=off} 
\floatstyle{ruled}
\newfloat{listing}{tb}{lst}{}
\floatname{listing}{Listing}
\usepackage{amsmath}
\usepackage{amsfonts}
\usepackage{graphicx}
\usepackage{subcaption}
\usepackage{tikz}
\usepackage{multirow}
\usepackage{booktabs}
\usepackage{tcolorbox}
\usepackage{multicol}
\usepackage{algorithm}
\usepackage{soul}
\usepackage{array}
\usepackage{arydshln}
\usetikzlibrary{bayesnet}

\title{Learning to Disentangle Latent Reasoning Rules with Language VAEs: \\ A Systematic Study}

\author{Yingji Zhang$^{1\dagger}$,~ Marco Valentino$^{2}$, ~ Danilo S. Carvalho$^{1,4}$, ~ Andr\'{e} Freitas$^{1,3,4}$ \\
$^{1}$ Department of Computer Science, University of Manchester, UK\\
$^{2}$School of Computer Science, University of Sheffield, UK\\
$^{3}$ Idiap Research Institute, Switzerland\\
$^{4}$ Cancer Biomarker Centre, CRUK Manchester Institute, UK\\
}

\begin{document}

\maketitle
\begin{abstract}
Incorporating explicit reasoning rules within the latent space of language models (LMs) offers a promising pathway to enhance generalisation, interpretability, and controllability. While current Transformer-based language models have shown strong performance on Natural Language Inference (NLI) tasks, they often rely on memorisation rather than explicit rule-based generalisation. This work investigates how human-interpretable reasoning rules can be explicitly encoded within
LMs with the support of Language Variational Autoencoders (VAEs), as a mechanism for generative control. We propose a complete pipeline for learning reasoning rules within Transformer-based language VAEs. This pipeline encompasses three rule-based reasoning tasks, a supporting theoretical framework, and a practical end-to-end architecture. The experiment illustrates the following findings: \textbf{Disentangled reasoning:} Under explicit signal supervision, reasoning rules (viewed as functional mappings) can be disentangled within the encoder’s parametric space. This separation results in distinct clustering of rules in the output feature space. \textbf{Prior knowledge injection:} injecting rule-based constraints into the Query enables the model to more effectively retrieve the stored Value from memory based on Key. This approach offers a simple method for integrating prior knowledge into decoder-only language models. 
Moreover, we found that FFN layers are better than attention layers at preserving the separation of reasoning rules in the model's parameters.
\end{abstract}
\section{Introduction}







\begin{figure*}[t]
    \centering
    \includegraphics[width=\linewidth]{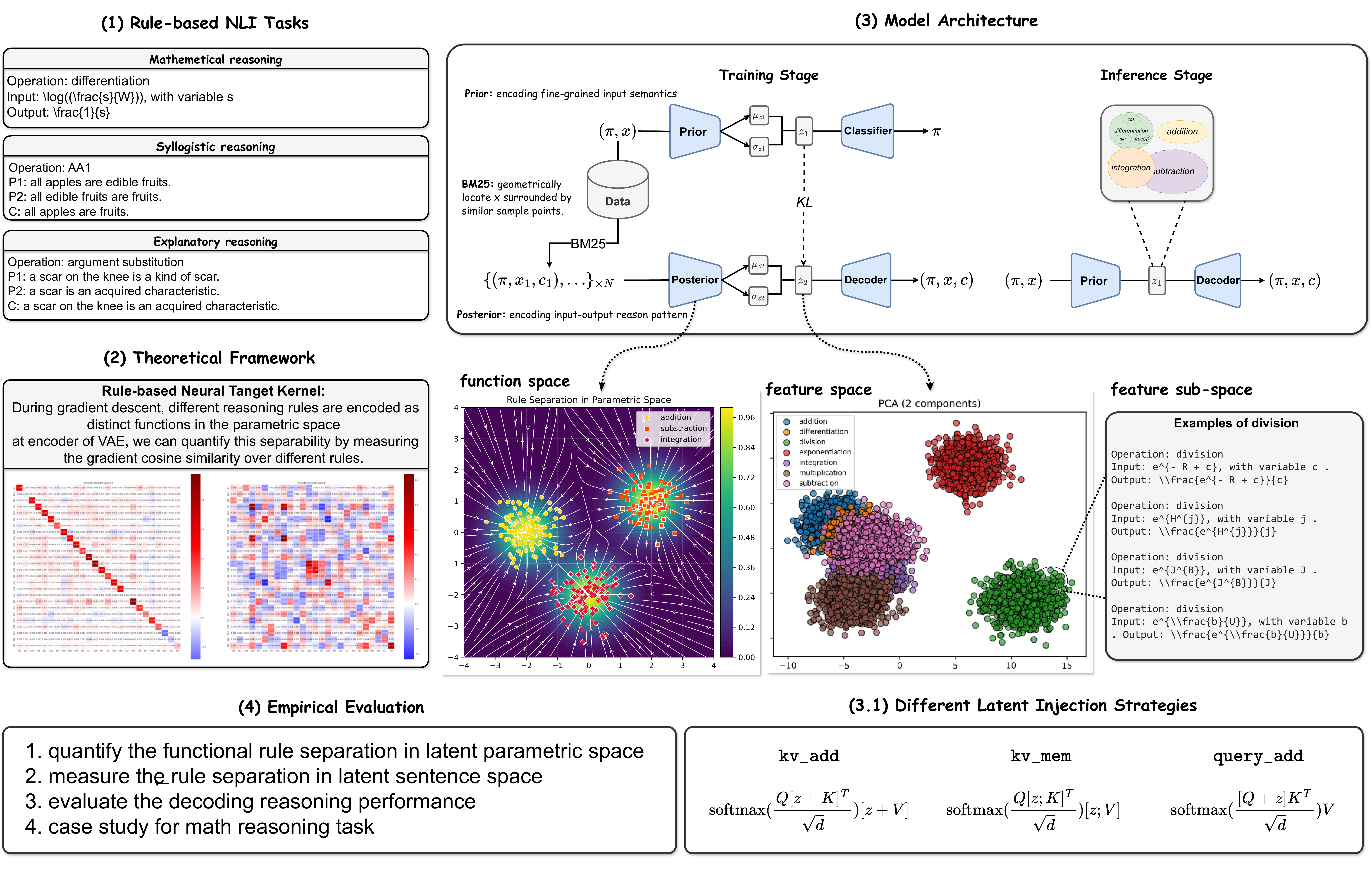}
    \caption{Overview, where $(\pi, x, c)$ represents the \textit{(rule, input premise(s), conclusion)}. To systematically evaluate rule-based learning within a VAE framework, first, we examine three rule-based NLI tasks. Second, we formalise the hypothesis that reasoning rules can be functionally and separately encoded within the encoder’s parametric space, enabling rule learning in the latent space, grounded in the theoretical framework of neural tangent kernels. Third, we introduce an end-to-end VAE architecture, with three different latent injection setups, designed to capture coarse-grained reasoning patterns in its latent space while remaining sensitive to the lexical semantics of the input.}
    \label{fig:method}
\end{figure*}

Encoding reasoning patterns as explicit rules within latent representations holds significant promise for enhancing the generalisation, interpretability, and controllability of neural models, with high downstream impact on AI safety and regulatory compliance \cite{bonnet2024searchinglatentprogramspaces,yu2025grammarbasedordinarydifferentialequation}. Over recent years, Transformer-based language models (LMs) have achieved notable success across a variety of Natural Language Inference (NLI) tasks \cite{yang2024qwen2technicalreport,qwen2025qwen25technicalreport}. Nonetheless, a growing body of research has demonstrated that in many instances these models often rely on memorisation rather than generalisation and that rule-based control mechanisms fail to be fully enforced \cite{yan2025phdlevelllmstrulygrasp}.

Therefore, this investigation focuses on the question: \textit{How can reasoning rules be explicitly encoded within the latent space of language models?} 
In the context of this work, a reasoning rule is defined as an input–output pattern that reflects a specific transformation of inference behaviour. It is important to note that this study does not aim to explore rule composition or generalisation. Rather, the primary objective is to explicitly encode reasoning input–output constraints within the model’s latent space, targeting better interpretability and controllability in the latent space.

Variational Autoencoders (VAEs) \cite{kingma2013auto} provide a compelling framework for this direction, where the integration of prior distribution serves as an inductive bias, enabling the model to leverage existing knowledge and providing a principled way to incorporate domain constraints \cite{papamarkou2024position}. This work focuses on lightweight LMs ($<1B$) as VAE decoders, allowing accessible evaluation of training dynamics, memory capacity, and model updates of the Transformer architecture \cite{zhong2025understanding,morris2025languagemodelsmemorize}. Additionally, because our primary motivation is to study latent rule learning within the encoder latent space, we fully re-train the decoder. As a result, the geometrical properties of the latent space are not constrained by the model’s architecture, enabling a more controlled analysis of representation learning.

Accordingly, we propose a complete pipeline for learning NLI reasoning rules within Transformer-based language VAEs \cite{li2020optimus,zhang-etal-2024-graph}: 

\textbf{First,} we focus on three distinct rule-based reasoning tasks, each characterised by unique syntax, inference patterns, and levels of granularity: Mathematical Reasoning \cite{meadows2023symbolic}, Syllogistic Reasoning \cite{valentino2025mitigatingcontenteffectsreasoning}, and Explanatory Reasoning \cite{zhang2024controllablenaturallanguageinference}.

\textbf{Second,} following the Neural Tangent Kernel (NTK) theory \cite{jacot2018ntk}, we claim that by supervising the inference rule information, inference rules can be encoded in the parametric space (i.e., weight matrix) of the encoder, which subsequently leads to the rule separation in the latent sentence space. This formalisation allows us to explicitly encode reasoning rules in the latent space.

\textbf{Third,} we propose an end-to-end VAE architecture where the latent space can encode both coarse-grained reasoning rules and fine-grained reasoning patterns that arise from semantic differences in the inputs, and evaluate three ways to inject the latent space into the LM decoder by intervening in the attention network.

We conduct extensive experiments to assess the effectiveness of rule encoding and reasoning generation capabilities which elicits the following results and findings:

\noindent\textbf{Disentangled Reasoning:} Under explicit signal supervision, reasoning rules, as functional mappings, can be disentangled within the encoder’s parametric space. This separation results in distinct clustering of rules in the output feature space, suggesting the potential for employing rule-based NTK theory to better understand the training dynamics and internal geometry of gradient-based neural NLI model.

\noindent\textbf{Prior Knowledge Injection:} Different latent space injection setups result in varying levels of reasoning performance. The optimal configuration is achieved by injecting the latent space into the Query of the attention network. Intuitively, injecting rule-based reasoning constraints into the Query enables the model to more effectively retrieve the stored Value from memory based on Key. This approach offers a simple method for integrating prior knowledge into LMs.

\noindent\textbf{Information Bottleneck:} A case study on the mathematical reasoning task reveals a performance bottleneck in decoder-only LMs, where increasing the number of samples per operation fails to yield performance improvements beyond a certain threshold. Additionally, we observe that FFN networks are more effective than attention in maintaining the separation of reasoning rules within the learned parametric space. Both provide additional insights on understanding Transformer architectures through memorisation.

To the best of our knowledge, this is the first study to explore the explicit encoding of NLI rules within language VAE latent spaces, targeting better latent space geometry to support rule-based inferences.

\section{Related Work} \label{sec:related}
In this section, we review related work around two topics: \textit{rule-based representation learning} and \textit{language VAEs}, to highlight current research limitations and elucidate the motivation underlying our work.

\paragraph{Rule-based Representation Learning.} Encoder-only models have been employed for tasks such as mathematical operations, where the encoder learns structured transformations, as demonstrated by \cite{valentino2024multioperationalmathematicalderivationslatent}. In addition, VAE-based approaches have shown promise in tasks requiring structured reasoning, such as program synthesis, where the goal is to generate programs that fulfil a specified task \cite{pmlr-v80-sun18a, bonnet2024searchinglatentprogramspaces, vankrieken2025neurosymbolicdiffusionmodels}. Similarly, grammar-based approaches using VAEs have been applied to infer ordinary differential equation (ODE) formulas from data \cite{yu2025grammarbasedordinarydifferentialequation}. Moreover, Decoder-only models, particularly large language models (LLMs), leverage in-context learning by using demonstrations to infer and apply underlying reasoning patterns \cite{liu2023context,bhattamishraunderstanding}. Despite these advancements, relatively few studies have investigated rule-based learning in the context of NLI, where this study focuses on. 

\paragraph{Language VAEs.} Language VAEs have been widely applied in NLP tasks, such as style transfer tasks: modifying sentences with regard to markers of sentiment, formality, affirmation/negation \cite{ shen2020educating,john2019disentangled,bao2019generating,hu2021causal,vasilakes-etal-2022-learning,gu-etal-2022-distributional,liu-etal-2023-composable,gu-etal-2023-controllable}, story generation \cite{fang2021transformerbased}, dialogue generation \cite{zhao-etal-2017-learning}, text paraphrasing \cite{bao-etal-2019-generating}, and textual, syntactic, semantic representation learning domain, such as syntax disentanglement \cite{mercatali-freitas-2021-disentangling-generative}, semantic-syntax separation \cite{zhang-etal-2024-graph}, semantic disentanglement \cite{silva-de-carvalho-etal-2023-learning,zhang-etal-2024-learning}, etc. Comparatively, in this work we focus on Natural Language Inference (NLI) with an emphasis on rule-based control. 

In the next section, we start by introducing the reasoning tasks and provide a formal illustration of rule-based learning through Neural Tangent Kernel theory.
\section{Rule-based Learning for NLI} \label{sec:rule_def}
\subsection{Natural Language Inference Rules}
We investigate three distinct types of reasoning tasks: mathematical derivation \cite{meadows-etal-2024-symbolic}, syllogistic reasoning \cite{valentino2025mitigatingcontenteffectsreasoning}, and explanatory reasoning \cite{zhang2024controllablenaturallanguageinference}. Specifically, \textbf{(1)} Mathematical reasoning: Mathematical expressions \cite{valentino2023multioperational,meadows2023symbolic} follow a well-defined syntactic structure and set of symbolic rules that are notoriously difficult for neural models. The dataset \cite{meadows2023symbolic} includes seven human-annotated symbolic rules, encompassing operations such as \textit{differentiation} and \textit{integration}. \textbf{(2)} Syllogistic reasoning: Syllogistic reasoning involves classical categorical logic, including four standard forms: Universal Affirmative (A) — “All A are B,” Universal Negative (E) — “No A are B,” Particular Affirmative (I) — “Some A are B,” and Particular Negative (O) — “Some A are not B.” The dataset encodes 24 valid syllogistic inference patterns, such as AA1 (Barbara). \textbf{(3)} Explanatory reasoning: Consists of material inferences with clearly defined explanatory sentence structures  \cite{jansen2018worldtree,valentino2022hybrid} providing a semantically challenging yet sufficiently well-scoped scenario to evaluate the syntactic and semantic organisation of the space. Based on the EntailmentBank corpus \cite{dalvi2021explaining}, \citet{zhang2024controllablenaturallanguageinference} we can define ten inference types based on these explanatory patterns, providing a diverse set of tasks which instantiate rule-based reasoning. A summarisation of each task and corpus is provided in Table \ref{tab:task} in the supplementary material.

\subsection{Rule-based Neural Tangent Kernel} \label{sec:ntk}

This work aims to encode latent reasoning rules within a low-dimensional latent space to guide the LM decoder’s reasoning generation, leveraging the VAE framework. We frame rule-based learning as learning the transformation from input premise(s) to output conclusion within the encoder. We formalise this framework by introducing a novel method grounded in the \textit{Neural Tangent Kernel (NTK) theory} \cite{jacot2018ntk}.

\paragraph{Latent subspace separation.}
Let $\mathcal{M}$ be a VAE model parameterised by $\theta = (\theta_{\text{enc}}, \theta_{\text{dec}})$. Suppose the encoder can represent a set of symbolic inference rules $\Pi = \{\pi_1, \pi_2, \dots, \pi_n\}$. Then, under supervised training with rule annotations:

\textit{\textbf{Proposition:} the encoder's parameters $\theta_{\text{enc}}$ induce a parametric structure in which each inference rule $\pi_i$ corresponds to a distinct subspace $S_{\pi_i} \subseteq \mathbb{R}^D$ of the encoder representation space.}
\begin{equation}
    \forall \pi_i, \pi_j \in \Pi,\, \pi_i \neq \pi_j \quad \Rightarrow \quad S_{\pi_i} \cap S_{\pi_j} \approx \emptyset
\end{equation}

This parametric separation (Figure~\ref{fig:heatmap}) directly leads to clearly delineated latent feature subspaces (Figure~\ref{fig:pca}), each uniquely encoding a symbolic inference rule.

\paragraph{Connection to NTK theory.}
NTK theory provides a rigorous theoretical framework for understanding neural network training dynamics by examining the kernel induced by gradients of the network's outputs with respect to its parameters. It suggests that during gradient descent, the network effectively learns linear approximations within distinct functional subspaces, each approximating discrete symbolic reasoning rules within the target reasoning task. Formally, let $f_{\text{encode}}$ be the encoder function such that:
\begin{equation}
    f_{\text{encode}}: (\pi, x, c) \mapsto \mathbf{z}_{(\pi, x, c)} \in \mathcal{Z} \subseteq \mathbb{R}^D
\end{equation}

where $x$ is the input premise(s), c is the conclusion, $\pi \in \Pi$ is the reasoning rule, and $\mathcal{Z}$ is the latent representation space of dimension $D$. 
Each rule $\pi$ is explicitly embedded as part of the model input. As a result, the model effectively learns a function $f_\theta(\pi, x, c)$. The function $f_\theta$ thus jointly depends on both the content of transformation from premise(s) to conclusion and the nature of the symbolic operation to be performed.
Within the NTK framework, the similarity between two input examples of the same inference type $\pi$ is captured by the NTK as follows:
\begin{equation}
\Theta_{\pi}(x, x') = \nabla_\theta f_\theta(x, \pi)^\top \nabla_\theta f_\theta(x', \pi)
\end{equation}
\noindent where $x$ represents $(x,c)$ pair for concision. $\nabla_\theta f_\theta(x, \pi)$ denotes the gradient of the model output with respect to its parameters, evaluated at the input $(x, \pi)$. This kernel quantifies how a parameter update from one input-output pair would affect another pair, conditioned on the shared rule.

According to the NTK theory, the evolution of the model’s predictions under gradient descent training can be described by a linear kernel regression in the RKHS (Reproducing Kernel Hilbert Space) associated with $\Theta_{\pi}$. Specifically, the prediction at time $t$, $f_t(x, \pi)$, evolves as:
\begin{equation}
f_t(x, \pi) = f_0(x, \pi) - \Theta_{\pi}(x, \cdot) \left[\Theta_{\pi} + \lambda I\right]^{-1} (f_0 - c)
\end{equation}
\noindent where $f_0(x, \pi)$ is the model's output at initialisation for each training input, $\lambda$ is a regularisation parameter, and $c$ is the vector of ground truth conclusions.

Crucially, this formulation implies that each rule $\pi$ induces a distinct kernel $\Theta_{\pi}$, which in turn defines a unique RKHS $\mathcal{H}_{\pi}$—that is, a function space within which the model's solutions for rule $\pi$ reside. As the $\pi$ is varied, the structure of the kernel and the corresponding function space changes, reflecting the distinct reasoning behaviours associated with different inference operations. Thus, the model encodes different symbolic inference patterns in distinct, kernel-induced subspaces.


For two different rules, $\pi_i \neq \pi_j$, we examine the relationship between their corresponding NTKs, $\Theta_{\pi_i}$ and $\Theta_{\pi_j}$. Specifically, we are interested in the interaction between the parameter gradients induced by inputs associated with different inference types. Given two data points $x$ and $x'$, possibly corresponding to different premise pairs, the NTK entry for each rule is:
\begin{equation}
\Theta_{\pi_i}(x, x') = \nabla_\theta f_\theta(x, \pi_i)^\top \nabla_\theta f_\theta(x', \pi_i)
\end{equation}
and
\begin{equation}
\Theta_{\pi_j}(x, x') = \nabla_\theta f_\theta(x, \pi_j)^\top \nabla_\theta f_\theta(x', \pi_j)
\end{equation}
When considering cross-rule similarities, we are interested in the inner product between the gradients for different rules.
\begin{equation}
G_{ij}(x, x') := \langle \nabla_\theta f_\theta(x, \pi_i), \nabla_\theta f_\theta(x', \pi_j) \rangle
\end{equation}
If the rules $\pi_i$ and $\pi_j$ encode fundamentally different reasoning operations (e.g., \textit{addition} vs. \textit{subtraction} in Math Derivation), the gradients with respect to $\theta$ for inputs labeled with $\pi_i$ and those labeled with $\pi_j$ will tend to point in different directions in parameter space of the encoder.

Under idealised training, where the data for each rule is sufficiently distinct and the network has enough capacity, the gradients for one rule will have minimal overlap with those of the other. This can be formalised by observing that:
\begin{equation} \label{eq9}
\langle \nabla_\theta f_\theta(x, \pi_i), \nabla_\theta f_\theta(x', \pi_j) \rangle \approx 0 \qquad \text{for}~ \pi_i \neq \pi_j
\end{equation}
This property implies that the parameter updates driven by examples from different rules are approximately orthogonal, meaning that training on one type will not interfere with or alter the function learned for the other type. In the language of NTK and kernel regression, this corresponds to the induced RKHS for each type, $\mathcal{H}{\pi_i}$ and $\mathcal{H}{\pi_j}$, being approximately disjoint:
\begin{equation}
\mathcal{H}{\pi_i} \cap \mathcal{H}{\pi_j} \approx \emptyset
\end{equation}

In the next section, we will use this foundation to introduce the proposed VAE architecture and its supporting optimisation function.


\begin{table*}[htbp]
\centering
\renewcommand{\arraystretch}{1}
\resizebox{\linewidth}{!}{
\begin{tabular}{cccccccc}
\toprule
\textbf{Base Model} &
\multicolumn{2}{c}{\textbf{Math Derivations}} & 
\multicolumn{2}{c}{\textbf{Math Derivations (OOD)}} &
\multicolumn{2}{c}{\textbf{Syllogistic Reasoning}} &
\textbf{Explanatory Reasoning} \\
&
bleu & acc & bleu & acc & bleu & acc & bleu \\
\midrule
GPT2-medium & 0.2019 & 0.0171 & 0.0206 & 0.0018 & 0.0108 & 0.0000 & 0.5947 \\
Llama3-1B & 0.3412 & 0.0257 & 0.0625 & 0.0114 & 0.3612 & 0.1200 & 0.3160 \\
Qwen2-0.5B & 0.4200 & 0.1171 & 0.0869 & 0.0128 & 0.8130 & 0.4600 & 0.5372 \\
Qwen2.5-0.5B & \textbf{\textcolor{black}{0.6260}} & \textbf{\textcolor{black}{0.3800}} & \textbf{\textcolor{black}{0.1293}} & \textbf{\textcolor{black}{0.0185}} & \textbf{\textcolor{black}{0.9014}} & \textbf{\textcolor{black}{0.7000}} & \textbf{\textcolor{black}{0.6566}} \\
\bottomrule
\end{tabular}
}
\caption{Quantitative evaluation for decoder-only models. We can observe that Qwen2.5-0.5B demonstrates strong performance across various baselines, making it a suitable choice as the decoder component in our VAE architecture.} \label{tab:decode_lm}
\end{table*}
\section{Approach} \label{sec:latent_props}
\paragraph{Latent space properties.} We posit that the latent space should satisfy two essential geometrical properties:

\textit{Property 1: Rule-level encoding.} The latent space must capture the transformation defined by the reasoning rule $\pi$, which maps an input $x$ to a conclusion $c$, i.e., $\pi: x \rightarrow c$.

\textit{Property 2: Semantic-level encoding.} The latent space should also encode the lexical semantics of $x$, accounting for fine-grained variations in reasoning patterns that arise from semantic differences in the inputs. For example, within the Math Derivation task, under \textit{differentiation}, different inputs yield distinct transformations: $\pi: x_1 \rightarrow c_1$: 8 \textbackslash sin\{(u)\} $\rightarrow$ 8 \textbackslash cos\{\(u\)\} and $\pi: x_2 \rightarrow c_2$: \textbackslash log\{\(K\)\} $\rightarrow$ \textbackslash frac\{1\}\{K\}. By ensuring both properties, the latent space can capture both coarse-grained and fine-grained reasoning patterns.

\paragraph{Architecture.} We adopt a Transformer-based VAE framework, employing two distinct encoders, both instantiated with BERT~\cite{devlin2019bert}, to model the prior and posterior Gaussian distributions. The posterior encoder is optimised to capture the reasoning rule transformation, thereby satisfying \textit{Property 1}. Simultaneously, the prior encoder serves as a regulariser, encouraging the posterior to encode fine-grained sub-rule variations grounded in the lexical semantics of the input, thereby satisfying \textit{Property 2}. 

Concretely, given a target $(\pi, x)$ pair, we first retrieve a set of semantic similar examples $\{(\pi, x_1, c_1), \dots, (\pi, x_N, c_N)\}$ using a retrieval function (e.g. BM25 and N=12). These examples are defined as inputs to the posterior encoder, which learns a latent representation of the reasoning transformation by averaging the latent vectors of the retrieved instances. Geometrically, averaging the latent sample vectors positions the target $x$ at the centroid of the surrounding samples within the latent space, which is naturally aligned with how information is encoded in the latent space \cite{zhang2024formalsemanticgeometrytransformerbased}. The decoder then uses this aggregated representation to generate the corresponding conclusion $c$ for the target triplet $(\pi, x, c)$. In parallel, the prior encoder processes the target $(\pi, x)$ pair to predict the rule $\pi$ within the latent space via a linear classifier. By minimising the Kullback–Leibler (KL) divergence between the posterior and prior distributions, the latent space is encouraged to satisfy both geometric properties.

\paragraph{Latent injection.} We adopt three strategies to inject the latent variable $z$ into the decoder (e.g., Qwen2.5~\cite{qwen2025qwen25technicalreport}). Each approach modifies the attention mechanism to incorporate information from the latent space.

\paragraph{1. \texttt{kv\_add}} Following~\cite{zhang-etal-2024-graph}, the latent vector $z$ is added to both the Key ($K$) and Value ($V$) matrices in the attention network:
$\text{softmax}\left(\frac{\text{Q}[\text{z} + \text{K}]^\top}{\sqrt{d}}\right)[\text{z} + \text{V}]$

\paragraph{2. \texttt{kv\_mem}} Following~\cite{li2020optimus}, the latent vector $z$ is concatenated to the Key and Value matrices:
$\text{softmax}\left(\frac{\text{Q}[\text{z}; \text{K}]^\top}{\sqrt{d}}\right)[\text{z}; \text{V}]$

\paragraph{3. \texttt{query\_add}} In this setup, the latent vector $z$ is added to the Query ($Q$) matrix:
$\text{softmax}\left(\frac{[\text{Q} + \text{z}] \text{K}^\top}{\sqrt{d}}\right)\text{V}$

Here, $Q$, $K$, and $V$ denote the query, key, and value matrices in the attention mechanism, each with dimensions $\mathbb{R}^{\text{dim} \times \text{seq}}$, where $\text{dim}$ is the attention dimensionality and $\text{seq}$ is the sequence length.

\paragraph{Optimisation.} Finally, the model can be trained end-to-end via the evidence lower bound (ELBO) on the log-likelihood of the data $x$ \cite{kingma2013auto}. To avoid the KL vanishing issue, we select the cyclical schedule to increase weights of KL $\beta$ from 0 to 1 \cite{fu-etal-2019-cyclical} and a KL thresholding scheme \cite{li-etal-2019-surprisingly} that chooses the maximum between KL and threshold $\lambda$. The final objective function can be described as follows:
\begin{equation}
\begin{aligned} \label{eq:elbo_loss}
\mathcal{L}_\text{VAE} = & \mathbb{E}_{q_\phi(z|\pi,x,c)} \Big[ \log p_{\theta} (\pi,x,c | z ) \Big]  \\
& - \beta \max \left[ \lambda , \text{KL} q_\phi(z|\pi,x,c) || p(z|\pi,x) \right ] \\
& + \texttt{cls\_weight} \times \mathcal{L}_\text{classifier}
\end{aligned}
\end{equation}
\noindent where $q_\phi$, $p$, and $p_{\theta}$ represent the posterior encoder, prior encoder, and decoder, respectively. \texttt{cls\_weight} controls the strength of the classification loss. In our experiments, \texttt{cls\_weight} is evaluated at 1.0, 0.5, and 0.1.
\section{Empirical Analysis} \label{sec:empirical}
\subsection{Decoding Evaluation}
\paragraph{Base model.} First, we evaluate the reasoning performance of different decoder-only models, including Qwen2 \cite{yang2024qwen2technicalreport}, Qwen2.5 \cite{qwen2025qwen25technicalreport}, GPT2 \cite{radford2019language}, and trimmed Llama3 \cite{grattafiori2024llama3herdmodels}. Due to the scale of the dataset and limitations in computational resources, we restrict our fine-tuning to smaller models with fewer than 1 billion parameters. During training, the input format of decoder is described as:
$\text{operation:}~\pi,~\text{premise:}~x,~\text{conclusion:}~c$. During inference, we omit $c$, allowing the model to generate the conclusion. All models are evaluated on the same testset. Further experimental details are provided in the supplementary material.

To evaluate generation quality, we employ both the BLEU score \cite{Papineni02bleu:a} and accuracy (acc) as performance metrics. For the Math Reasoning task, in addition to the in-distribution test set, we also assess model performance on an out-of-distribution (OOD) test set, where the mathematical expressions are composed using a different set of variables.
\begin{table*}[htbp]
\centering
\resizebox{\linewidth}{!}{
\begin{tabular}{ccccccccc}
\toprule
\textbf{Injection} & \textbf{Train Setup} &
\multicolumn{2}{c}{\textbf{Math Reason}} & 
\multicolumn{2}{c}{\textbf{Math Reason (OOD)}} &
\multicolumn{2}{c}{\textbf{Syllogistic Reason}} &
\textbf{Explanatory Reason} \\
& &
bleu & acc & bleu & acc & bleu & acc & bleu \\
\midrule

\multirow{3}{*}{Base model}
    & Zero-shot & 0.6260 & 0.3800 & \textbf{0.1293} & 0.0185 & 0.9014 & 0.7000 & \underline{\textbf{0.6566}} \\
    & Few-shot & 0.1596 & 0.0614 & 0.1011 & 0.0057 & 0.1534 & 0.0000 & 0.0765 \\
    & Few-shot (FT) & 0.1601 & 0.0557 & 0.1022 & 0.0028 & 0.1712 & 0.0000 & 0.0801 \\
    \midrule
    \midrule
\multirow{3}{*}{kv\_add} 
    & prior=False & \textcolor{black}{\textbf{0.7285}} & \textbf{0.5285} & 0.1055 & 0.0200 & 0.4839 & 0.3400 & 0.6127 \\
    & weight=1.0 & 0.4986 & 0.2128 & 0.0869 & 0.0157 & 0.0719 & 0.0000 & 0.2049 \\
    & weight=0.5 & 0.4925 & 0.2814 & 0.0504 & 0.0185 & 0.4870 & 0.0900 & 0.2623 \\
\midrule
\multirow{3}{*}{kv\_mem} 
    & prior=False & 0.6430 & 0.4185 & 0.0985 & 0.0200 & 0.4672 & 0.3300 & 0.6313 \\
    & weight=1.0 & 0.5255 & 0.2300 & 0.0956 & 0.0185 & 0.0000 & 0.0000 & 0.6022 \\
    & weight=0.5 & 0.6808 & 0.4857 & 0.1130 & 0.0185 & 0.9452 & 0.8500 & 0.5987 \\
\midrule
\multirow{3}{*}{query\_add}
    & prior=False & 0.6501 & 0.3885 & 0.1130 & \textbf{0.0200} & \textbf{\underline{0.9681}} & \textbf{\underline{0.9200}} & \textbf{0.6449} \\
    & weight=1.0 & 0.7262 & 0.5057 & 0.1223 & \underline{\textbf{0.0214}} & 0.4373 & 0.3300 & 0.6118 \\
    & weight=0.5 & \underline{\textbf{0.8130}} & \underline{\textbf{0.6642}} & \underline{\textbf{0.1441}} & 0.0185 & \textbf{0.9461} & \textbf{0.8600} & 0.6220 \\
\bottomrule
\end{tabular}
}
\caption{Quantitative evaluation, where the base model is Qwen2.5-0.5B. Top two values are highlighed in \textbf{\underline{bold}} and \textbf{\textcolor{black}{bold}}. Prior=False is the setup of VAE without trainable prior. We can observe that query\_add leads to the best performance in general.} \label{tab:decode}
\end{table*}
As shown in Table \ref{tab:decode_lm}, Qwen2.5 demonstrates consistently strong performance across all tasks. Based on these results, we select Qwen2.5 as the decoder model for subsequent experiments.

\paragraph{VAE model.} Next, we evaluate the performance of the proposed VAE setting on downstream reasoning tasks, where the decoder is Qwen2.5-0.5B, the latent dimension is 32 following the same setup as Optimus \cite{li2020optimus}. In addition, we include results from the base model trained via fine-tuning (denoted as FT) as well as inference using few-shot examples. The input format remains consistent with previous settings, with examples repeated accordingly. For few-shot selection, we employ BM25 to retrieve the most relevant samples.

As illustrated in Table \ref{tab:decode}, we can observe that \ul{injecting the latent space into the query can generally result in better performance compared with base model and other setups}. Intuitively, injecting reasoning information into the Query enables the model to more effectively retrieve the stored Value from memory based on the Key \textbf{(Finding 1)}. In this setup, models with a trainable prior demonstrate improved performance on mathematical reasoning tasks, but not necessarily on other NLI tasks. This performance gain is attributed to the highly regular and syntactically consistent nature of the target mathematical expressions. In such contexts, retrieving structurally similar examples can effectively enhance model performance by reinforcing pattern recognition.

Furthermore, \ul{when the base model is trained using few-shot examples, its performance declines significantly. We observe that the model repeats to generate more examples}. To ensure a fair comparison, we did not apply any filtering to remove these redundant outputs \textbf{(Finding 2)}.

Additionally, we evaluate the performance of the VAE model trained with and without BM25-based sample selection. In the absence of the relevance function, training samples are randomly drawn from the corpus. In Table \ref{tab:decode_bm25}, \ul{integrating BM25-based selection during training yields enhanced performance, indicating the effectiveness of relevance-guided sampling}. From a geometric perspective, \ul{BM25 retrieves samples that exhibit the highest lexical similarity, effectively selecting the nearest neighbours in the latent space} \textbf{(Finding 3)}.
\begin{table}[ht]
\centering
\small
\begin{tabular}{ccccc}
\hline
\textbf{Injection} & \textbf{Prior} & \textbf{BM25} & \textbf{bleu} & \textbf{acc} \\ \hline
kv\_add & False & False & 0.3691 & 0.0614 \\
kv\_mem & False & False & 0.5525 & 0.2371 \\
query\_add & False & False & 0.5914 & 0.3021 \\ \hline
kv\_add & False & True & 0.7285 & 0.5285 \\
kv\_mem & False & True & 0.6430 & 0.4185 \\
query\_add & False & True & 0.6501 & 0.3885 \\ \hline
\end{tabular}
\caption{Quantitative evaluation for VAE model with or without BM25 in Math Reasoning task.} \label{tab:decode_bm25}
\end{table}
\begin{figure*}[ht!]
    \centering
    \includegraphics[width=0.8\linewidth]{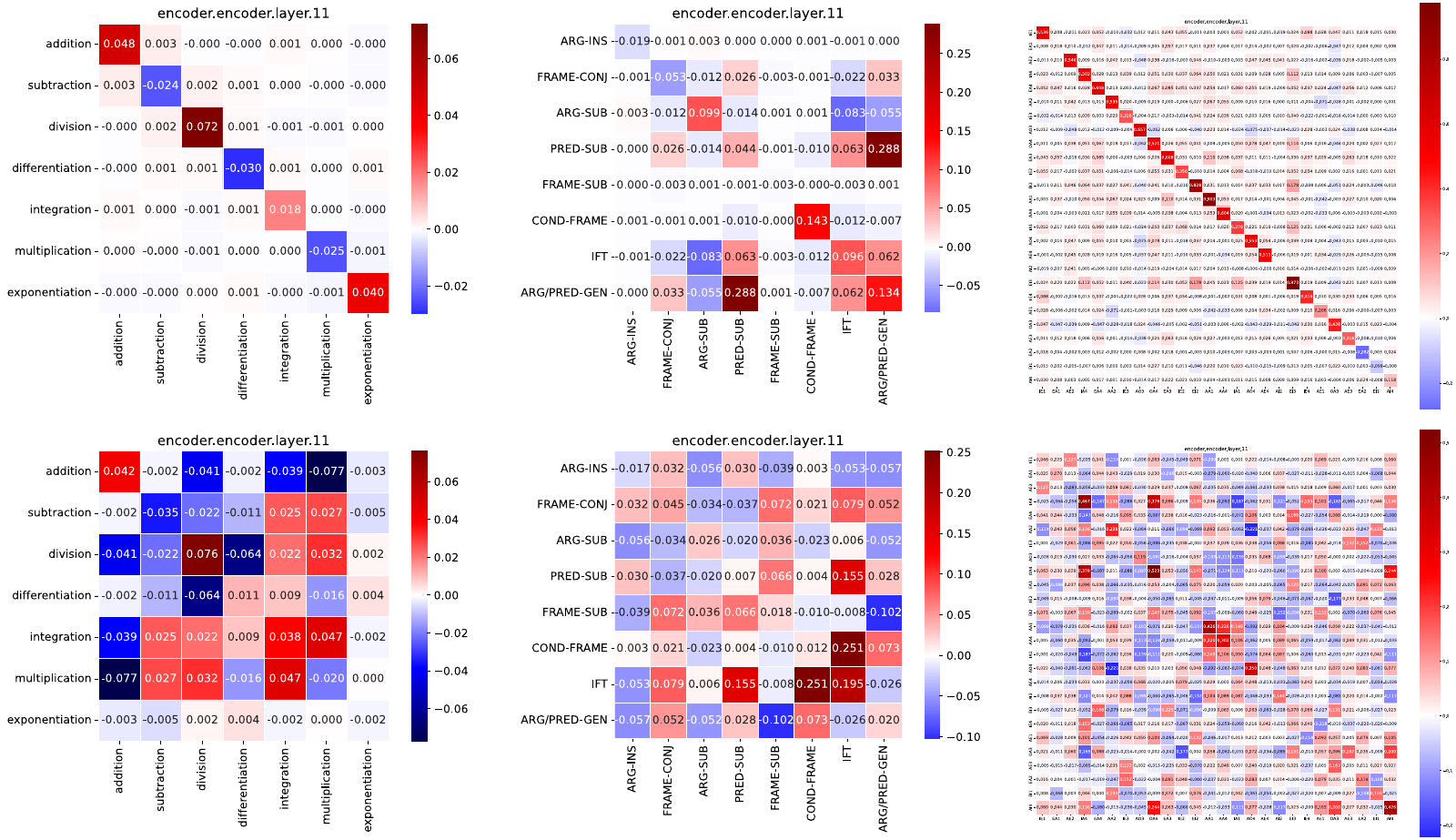}
    \caption{Gradient heatmap for the last posterior encoder layer (query\_add setup), where the left: math derivation, middle: explanatory reasoning, right: syllogistic reasoning, Top: \texttt{cls\_weight} is 1.0, bottom: \texttt{cls\_weight} is 0.1. We can observe that the non-diagonal values are notably close to 0 when providing higher \texttt{cls\_weight} (the red colour elements are less scattered), suggesting that incorporating rule information during training enhances the separation of rule subspaces in the encoder's parameter space. We provide the heatmaps of all layers in the supplementary material.}
    \label{fig:heatmap}
\end{figure*}
\begin{figure*}[ht!]
    \centering
    \includegraphics[width=\linewidth]{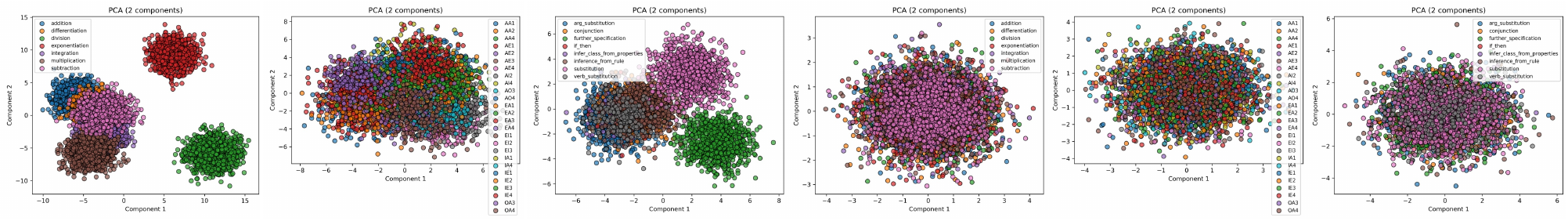}
    \caption{PCA visualisation for \texttt{query\_add} injection setup, where left three: \texttt{cls\_weight} is 1.0, right three: \texttt{cls\_weight} is 0.1. We can observe that the model struggle to learn the rules when the weight is close to zero, indicating the neural network try to deliver reason behaviour via memorisation, rather than rule-based learning. For other injection setups, their visualisations are provided in Figure \ref{fig:math_pca}, \ref{fig:explanation_pca}, and \ref{fig:syllogistic_pca} in the supplementary material.}
    \label{fig:pca}
\end{figure*}
\subsection{Encoding Evaluation}
\paragraph{Latent parametric space.} As illustrated in Section \ref{sec:ntk} (Equation \ref{eq9}), by measuring the cosine similarity between gradient vectors associated with different rules, we can quantify the separability between different rule subspaces in the encoder, comparing settings with strong \texttt{cls\_weight} = 1.0 (left) and weak \texttt{cls\_weight} =0.1 (right) during training. As illustrated in Figure \ref{fig:heatmap}, when the classification weight (\texttt{cls\_weight}) is set to 1.0, most non-diagonal values are close to zero (orthogonality). In contrast, with \texttt{cls\_weight} set to 0.1, a greater number of non-diagonal elements exhibit higher values (the red colour elements are much more scattered). This observation suggests that \ul{explicit supervision facilitates the separation of reasoning rules within the encoder’s parameter space} \textbf{(Finding 4)}.

\paragraph{Latent sentence space.} Since rule separation in the parametric space leads to corresponding separation in the feature space, as shown in Figure \ref{fig:pca}, the sentence representations tend to form distinct clusters that reflect rule information when the classifier is given a higher \texttt{cls\_weight}. However, \ul{when the \texttt{cls\_weight} approaches zero, these rule-based clusters disappear}. This \ul{suggests that the neural network relies more on the memorisation of lexical combination than on rule-based learning} \textbf{(Finding 5)}.
\begin{figure*}[ht!]
    \centering
    \includegraphics[width=\linewidth]{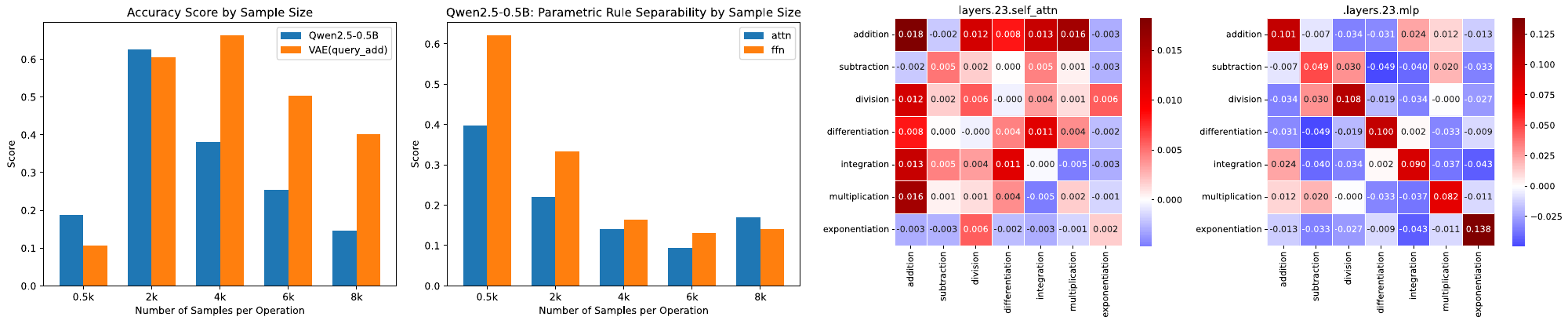}
    \caption{Case study for Math Reasoning task, where left: analysing how varying the number of training samples for each operation affects the reasoning capabilities. Right: comparing the parametric rule separation between attn and ffn at the last layer in Qwen2.5-0.5B, a pretrained checkpoint with a training sample size of 4k.}
    \label{fig:cases}
\end{figure*}
\subsection{Case Study for Math Reasoning}

\paragraph{Information bottleneck.} First, we analyse how varying the number of training samples for each operation affects the reasoning capabilities of the decoder-only LM (Qwen2.5-0.5B). In Figure \ref{fig:cases} (left bar plot), it can be observed that when the number of samples per category exceeds 2,000, there is a noticeable decline in accuracy. This suggests that increasing the sample size may introduce greater variability or complexity, potentially disrupting the consistency of each operation. \ul{The observation also highlights a limitation of current autoregressive LMs: rather than engaging in rule-based reasoning, they tend to rely on retrieving memorised training instances embedded in their parameters} \cite{zhong2025understanding}. Explicitly injecting latent reasoning representations can help mitigate this issue \textbf{(Finding 6)}.

\paragraph{Parametric rule separation.} Second, we assess the parametric separation of rule-based information across different model components, specifically the attention (attn) and feed-forward network (FFN) layers, using the same methodology outlined previously. We report the sum of the average diagonal values across all layers. As illustrated in Figure \ref{fig:cases} (right bar plot), \ul{the FFN layers (right) exhibit a greater tendency to encode rule-based information compared to the attention layers, indicating a more prominent role in capturing structured reasoning patterns} \textbf{(Finding 7)}.

\section{Conclusion and Future Work} \label{sec:concl}
This work serves as a foundational step in exploring rule-based representation learning under the language VAE architecture for NLI tasks. We propose a complete pipeline for learning reasoning rules within Transformer-based language VAEs. This pipeline encompasses three rule-based reasoning tasks, a supporting theoretical framework, and a practical end-to-end architecture. The experiment illustrates the following findings: \textbf{Disentangled reasoning:} Under explicit signal supervision, reasoning rules can be disentangled within the encoder’s parametric space, resulting in distinct clustering of rules in the output feature space. \textbf{Prior knowledge injection:} injecting reasoning information into the Query enables the model to more effectively retrieve the stored Value from memory based on Key. This approach offers a simple method for integrating prior knowledge into decoder-only language models. In addition, we found that FFN layers are better than attention layers at preserving the separation of reasoning rules in the model's parameters. In the future, we will focus on the investigation of diffusion-based language models, such as Mmada \cite{yang2025mmadamultimodallargediffusion}, which can improve flexibility for constraining the decoding process.

\bibliography{tacl2021}

\section*{Reproducibility Checklist}

Unless specified otherwise, please answer “yes” to each question if the relevant information is described either in the paper itself or in a technical appendix with an explicit reference from the main paper. If you wish to explain an answer further, please do so in a section titled “Reproducibility Checklist” at the end of the technical appendix. This paper:

\begin{enumerate}
\item Includes a conceptual outline and/or pseudocode description of AI methods introduced (yes)
\item Clearly delineates statements that are opinions, hypothesis, and speculation from objective facts and results (yes)
\item Provides well marked pedagogical references for less-familiare readers to gain background necessary to replicate the paper (yes)
\item Does this paper make theoretical contributions? (yes)
\end{enumerate}

If yes, please complete the list below.

\begin{enumerate}
    \item All assumptions and restrictions are stated clearly and formally. (yes)
    \item All novel claims are stated formally (e.g., in theorem statements). (yes)
    \item Proofs of all novel claims are included. (yes)
    \item Proof sketches or intuitions are given for complex and/or novel results. (yes)
    \item Appropriate citations to theoretical tools used are given. (yes)
    \item All theoretical claims are demonstrated empirically to hold. (yes)
    \item All experimental code used to eliminate or disprove claims is included. (yes)
    \item Does this paper rely on one or more datasets? (yes)
\end{enumerate}

If yes, please complete the list below.

\begin{enumerate}
    \item A motivation is given for why the experiments are conducted on the selected datasets (yes)
    \item All novel datasets introduced in this paper are included in a data appendix. (yes)
    \item All novel datasets introduced in this paper will be made publicly available upon publication of the paper with a license that allows free usage for research purposes. (yes)
    \item All datasets drawn from the existing literature (potentially including authors’ own previously published work) are accompanied by appropriate citations. (yes)
    \item All datasets drawn from the existing literature (potentially including authors’ own previously published work) are publicly available. (yes)
    \item All datasets that are not publicly available are described in detail, with explanation why publicly available alternatives are not scientifically satisficing. (yes)
    \item Does this paper include computational experiments? (yes)
\end{enumerate}

If yes, please complete the list below.

\begin{enumerate}
\item This paper states the number and range of values tried per (hyper-) parameter during development of the paper, along with the criterion used for selecting the final parameter setting. (yes)

\item Any code required for pre-processing data is included in the appendix. (yes).

\item All source code required for conducting and analyzing the experiments is included in a code appendix. (yes)

\item All source code required for conducting and analyzing the experiments will be made publicly available upon publication of the paper with a license that allows free usage for research purposes. (yes)

\item All source code implementing new methods have comments detailing the implementation, with references to the paper where each step comes from (yes)

\item If an algorithm depends on randomness, then the method used for setting seeds is described in a way sufficient to allow replication of results. (yes)

\item This paper specifies the computing infrastructure used for running experiments (hardware and software), including GPU/CPU models; amount of memory; operating system; names and versions of relevant software libraries and frameworks. (yes)

\item This paper formally describes evaluation metrics used and explains the motivation for choosing these metrics. (yes)

\item This paper states the number of algorithm runs used to compute each reported result. (yes)

\item Analysis of experiments goes beyond single-dimensional summaries of performance (e.g., average; median) to include measures of variation, confidence, or other distributional information. (yes)

\item The significance of any improvement or decrease in performance is judged using appropriate statistical tests (e.g., Wilcoxon signed-rank). (yes)

\item This paper lists all final (hyper-)parameters used for each model/algorithm in the paper’s experiments. (yes)

\end{enumerate}
\clearpage
\appendix

\section{Experimental Details}
\paragraph{Corpus.} Table \ref{tab:task} describes the reasoning tasks in the experiment. For each corpus, the dataset is split into train $60\%$, valid $20\%$, and test $20\%$ sets.

\paragraph{Train setup.} Max epoch: 10, lr: 3e-5, GPU device: A6000 $\times$ 2. The latent dimension is 32 during the experiment following the same setup as Optimus. All other experimental details are provided in the codebase.
\begin{table*}[h!]
\centering
\small
\begin{tabular}{@{}llcc@{}}
\toprule
\textbf{Task} & \textbf{Example} & \textbf{Num of Rules} & \textbf{Sizes} \\
\midrule
Math Reason & 
\begin{tabular}[c]{@{}l@{}}Operation: differentiation\\ Input: \textbackslash log((\textbackslash frac\{s\}\{W\})), with variable s\\ Output: \textbackslash frac\{1\}\{s\}
\end{tabular} & 
7 & 28k \\
\midrule
Explanatory Reason & 
\begin{tabular}[c]{@{}l@{}}Operation: argument substitution\\ P1: a scar on the knee is a kind of scar.\\ P2: a scar is an acquired characteristic.\\ C: a scar on the knee is an acquired characteristic.
\end{tabular} & 
8 & \textasciitilde5k \\
\midrule
Syllogistic Reason & 
\begin{tabular}[c]{@{}l@{}}Operation: AA1\\ P1: all apples are edible fruits.\\ P2: all edible fruits are fruits.\\ C: all apples are fruits.
\end{tabular} & 
26 & \textasciitilde5k \\ \bottomrule
\end{tabular}
\caption{Reasoning tasks, examples, number of rules, and dataset sizes.} \label{tab:task}
\end{table*}
\section{Additional Results} \label{sec:addition}

\begin{figure*}[ht!]
    \centering
    \includegraphics[width=\linewidth]{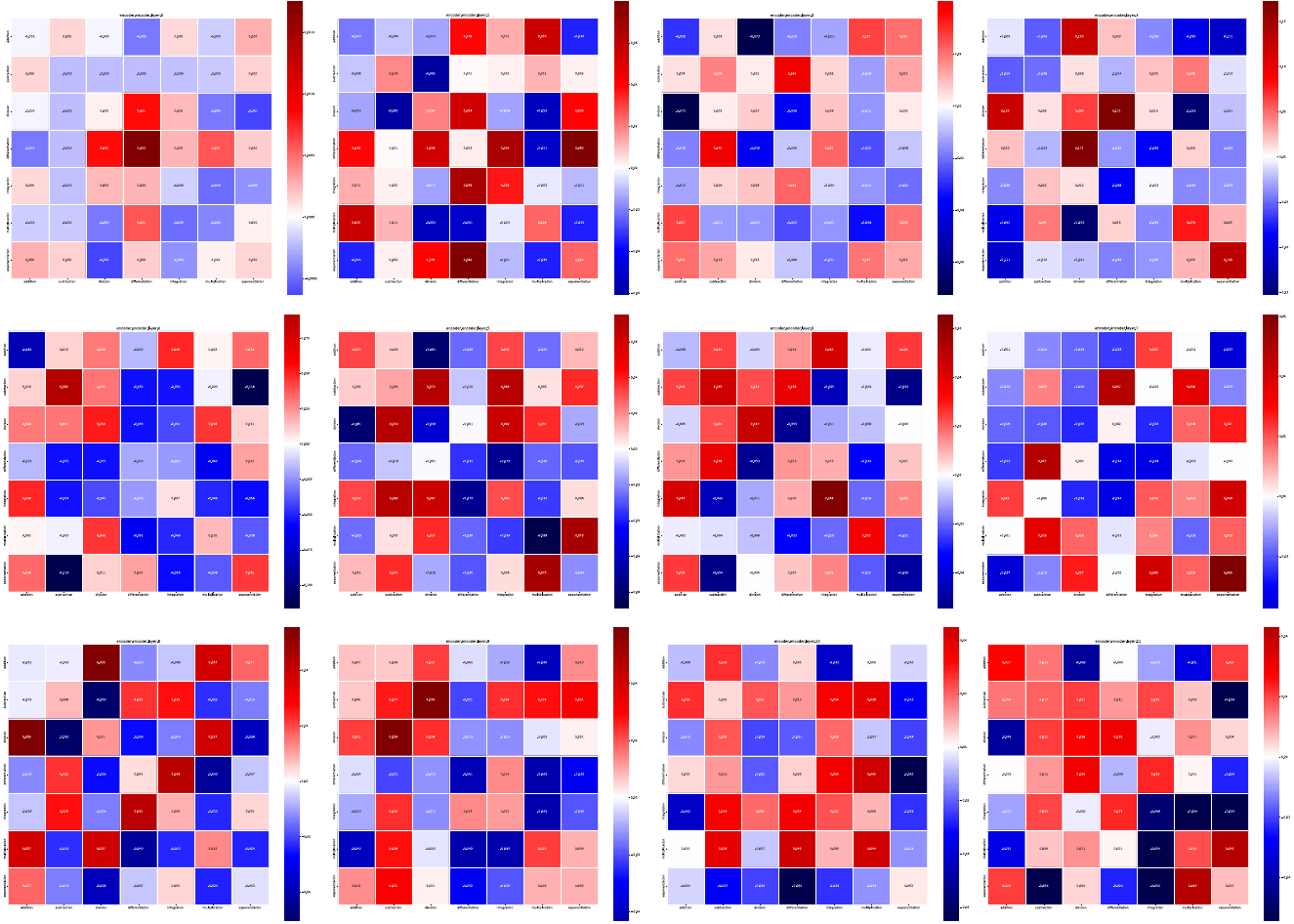}
    \caption{Math Reasoning: gradient heatmap, where cls\_weight is 0.1.}
    \label{fig:grad_math_0}
\end{figure*}

\begin{figure*}[ht!]
    \centering
    \includegraphics[width=\linewidth]{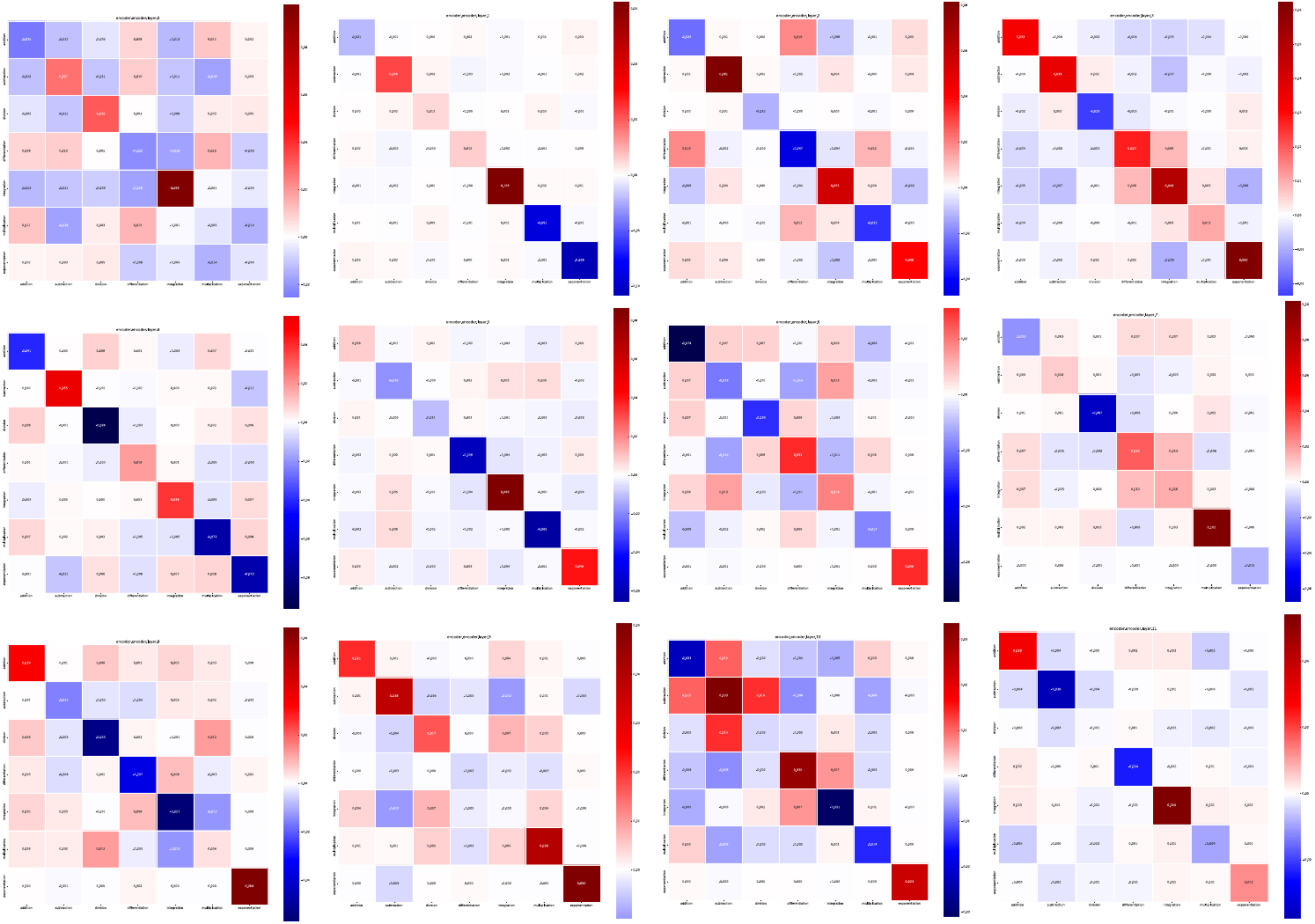}
    \caption{Math Reasoning: gradient heatmap, where cls\_weight is 1.0.}
    \label{fig:grad_math_1}
\end{figure*}

\begin{figure*}[ht!]
    \centering
    \includegraphics[width=\linewidth]{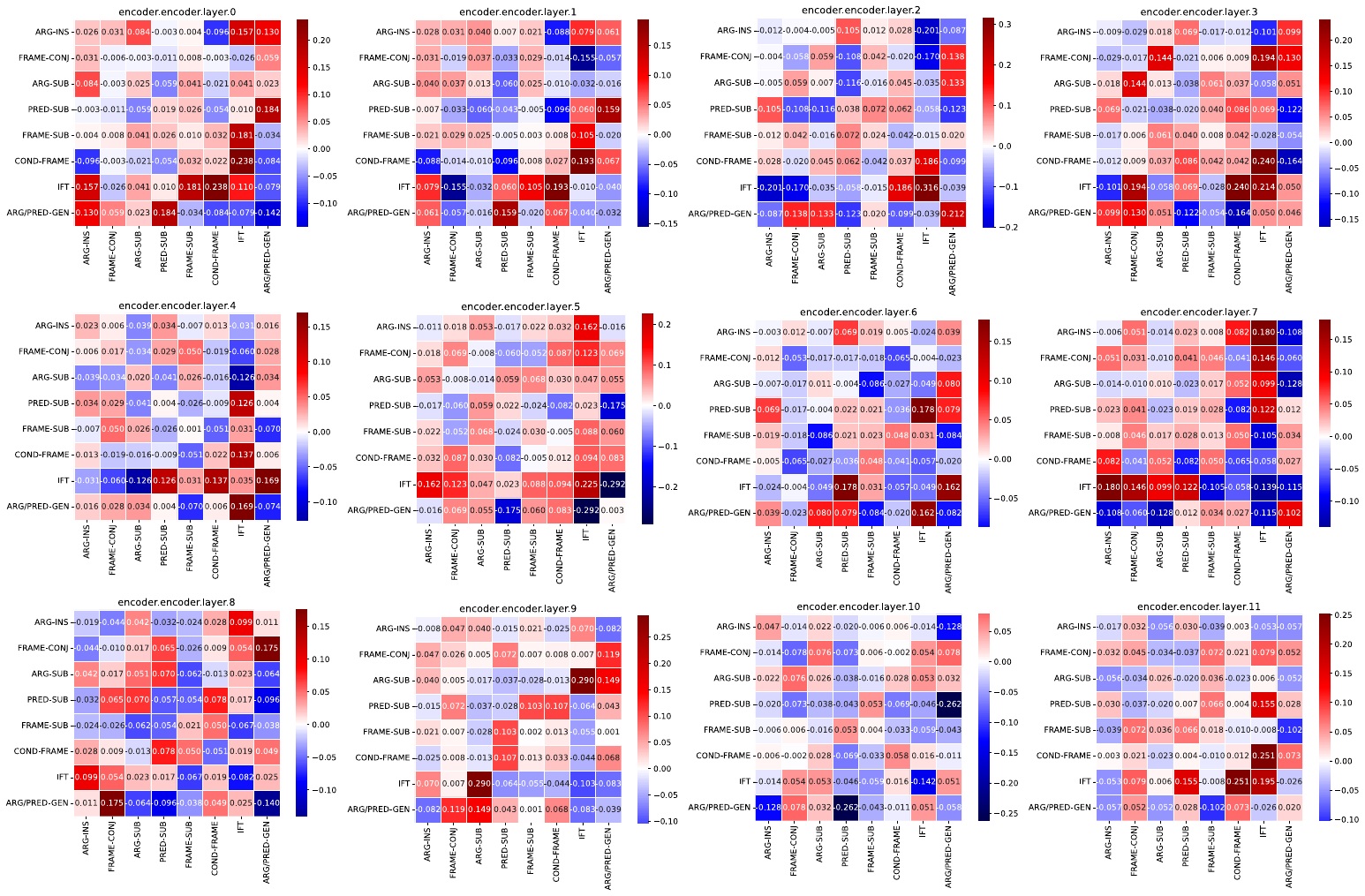}
    \caption{Explanatory Reasoning: gradient heatmap, where cls\_weight is 0.1.}
    \label{fig:grad_exp_0}
\end{figure*}

\begin{figure*}[ht!]
    \centering
    \includegraphics[width=\linewidth]{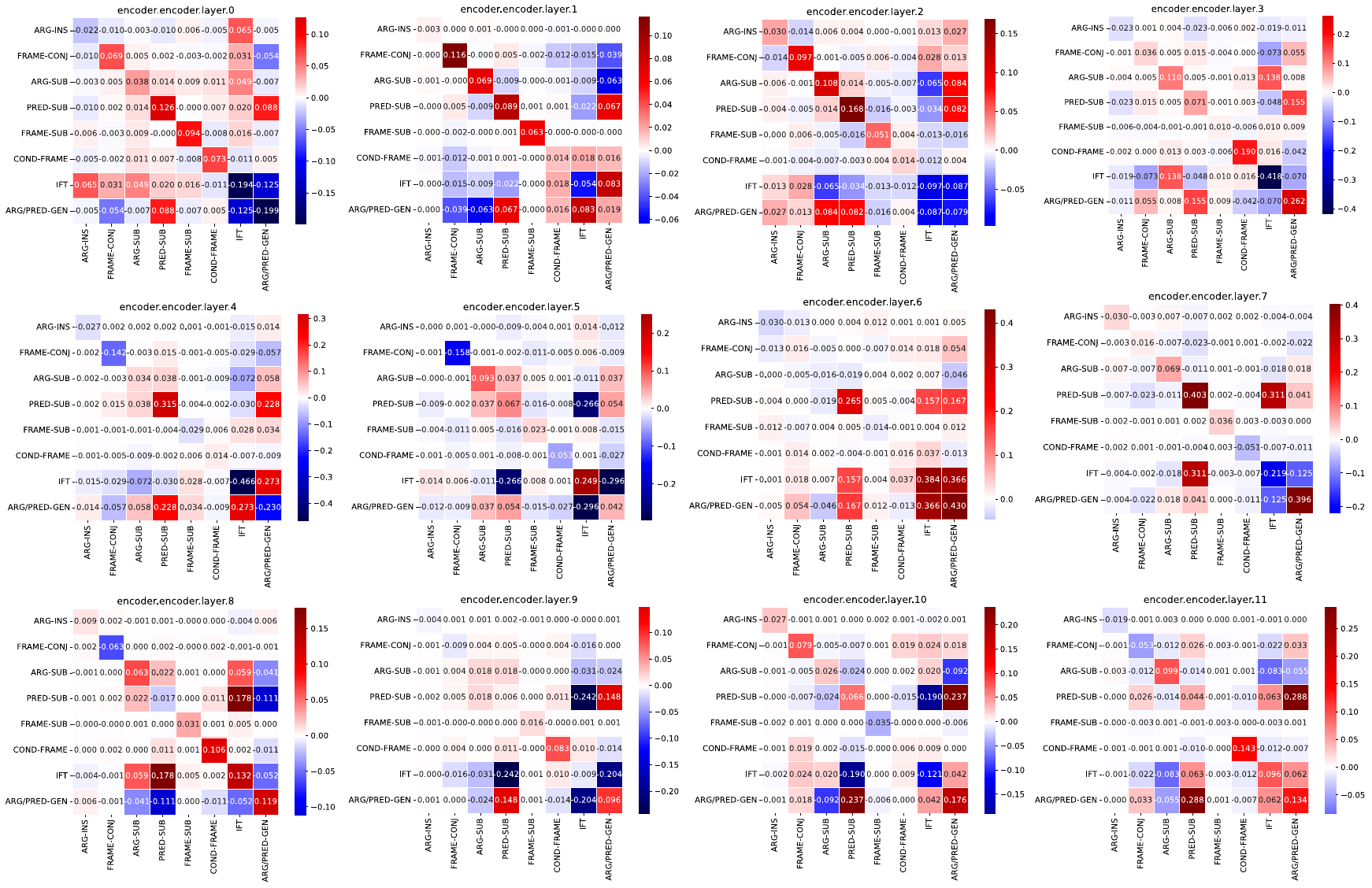}
    \caption{Explanatory Reasoning: gradient heatmap, where cls\_weight is 1.0.}
    \label{fig:grad_exp_1}
\end{figure*}

\begin{figure*}[ht!]
    \centering
    \includegraphics[width=\linewidth]{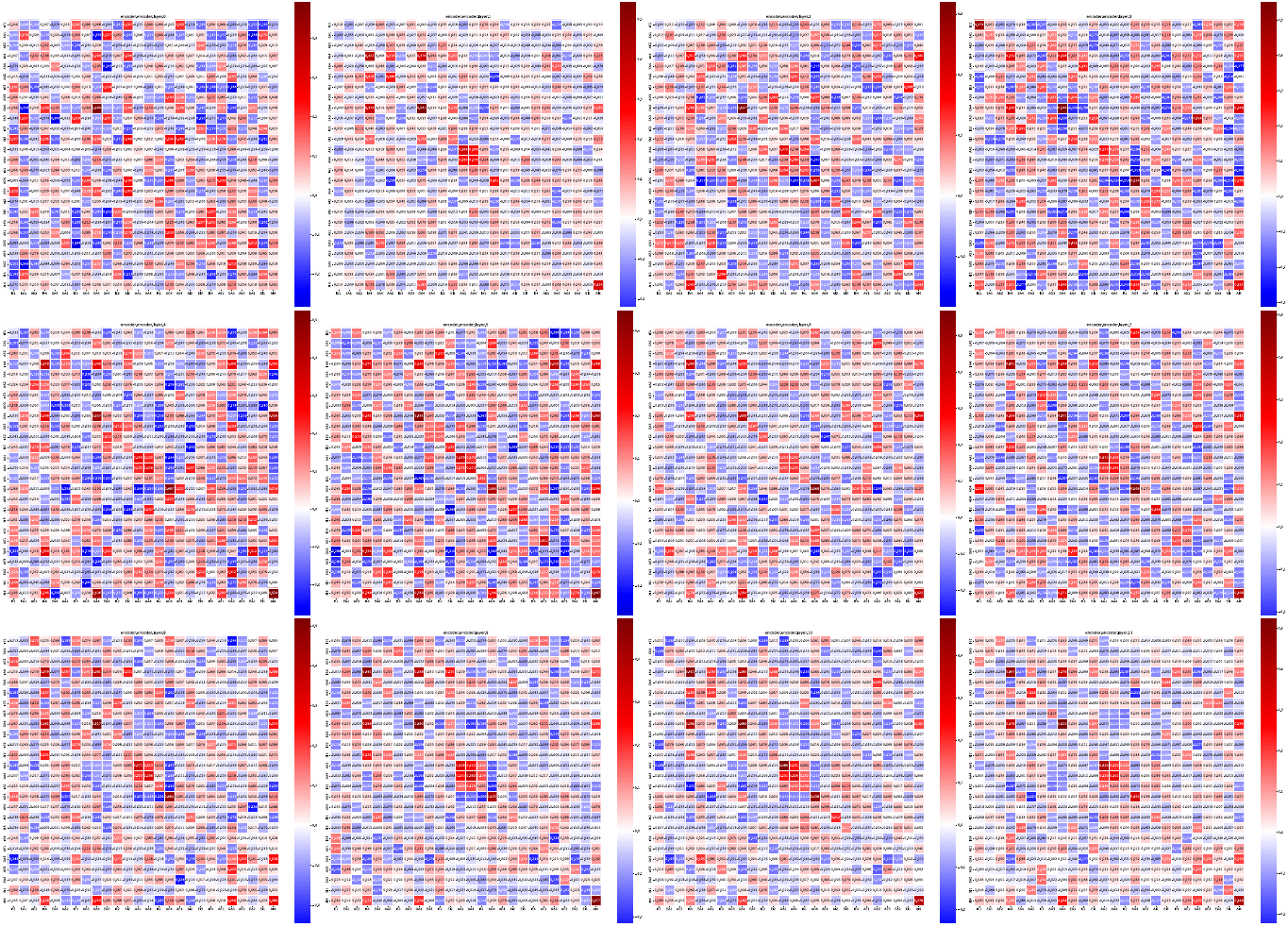}
    \caption{Syllogistic Reasoning: gradient heatmap, where cls\_weight is 0.1.}
    \label{fig:grad_syl_0}
\end{figure*}

\begin{figure*}[ht!]
    \centering
    \includegraphics[width=\linewidth]{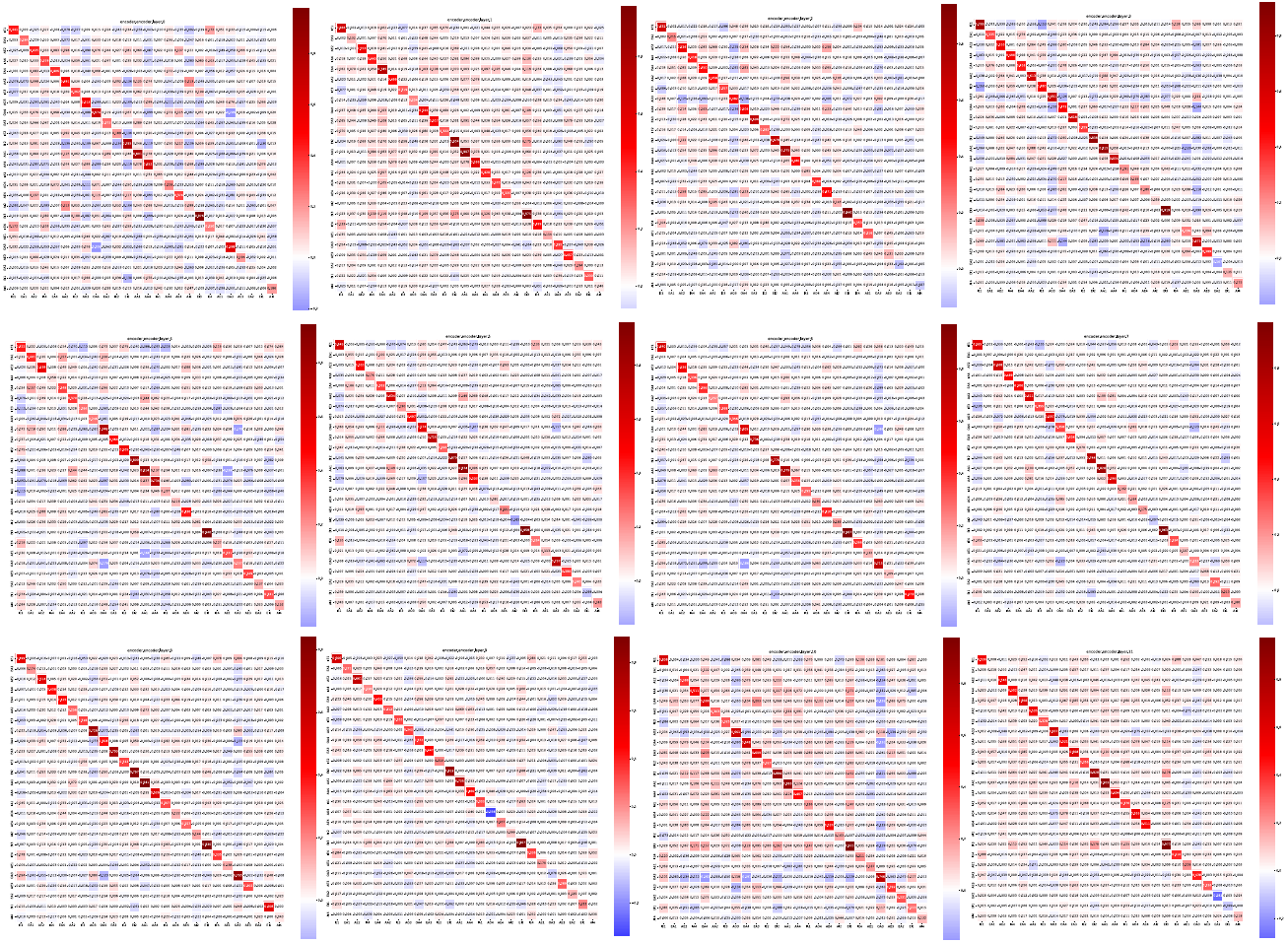}
    \caption{Syllogistic Reasoning: gradient heatmap, where cls\_weight is 1.0.}
    \label{fig:grad_syl_1}
\end{figure*}

\paragraph{Parametric space visualisation.} We provide comprehensive heatmaps covering all encoder layers across the full set of tasks, including Math Reasoning: Figure \ref{fig:grad_math_0} and \ref{fig:grad_math_1}; Explanatory Reasoning: Figure \ref{fig:grad_exp_0} and \ref{fig:grad_exp_1}; Syllogistic Reasoning: Figure \ref{fig:grad_syl_0} and \ref{fig:grad_syl_1}.

\begin{figure*}[ht!]
    \centering
    \includegraphics[width=\linewidth]{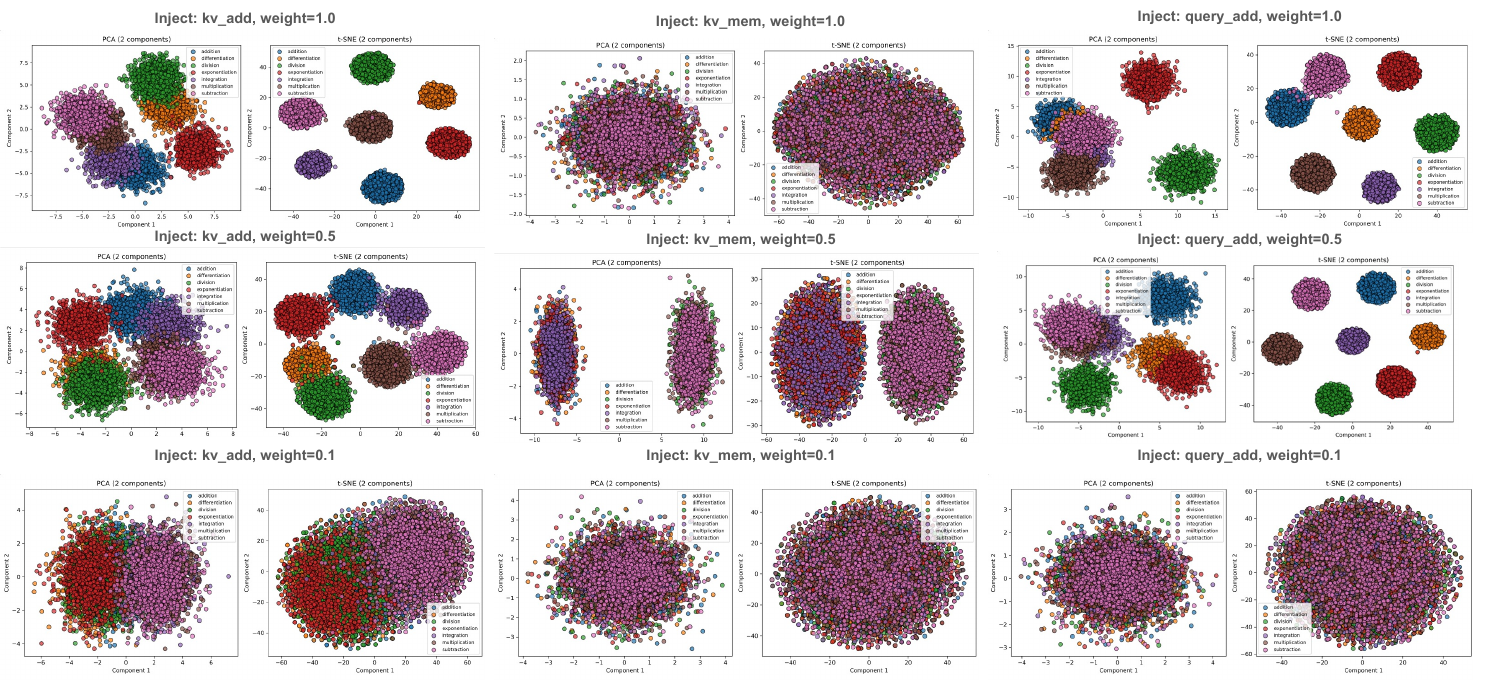}
    \caption{PCA and T-SNE visualisation for Math Derivation task.}
    \label{fig:math_pca}
\end{figure*}

\begin{figure*}[ht!]
    \centering
    \includegraphics[width=\linewidth]{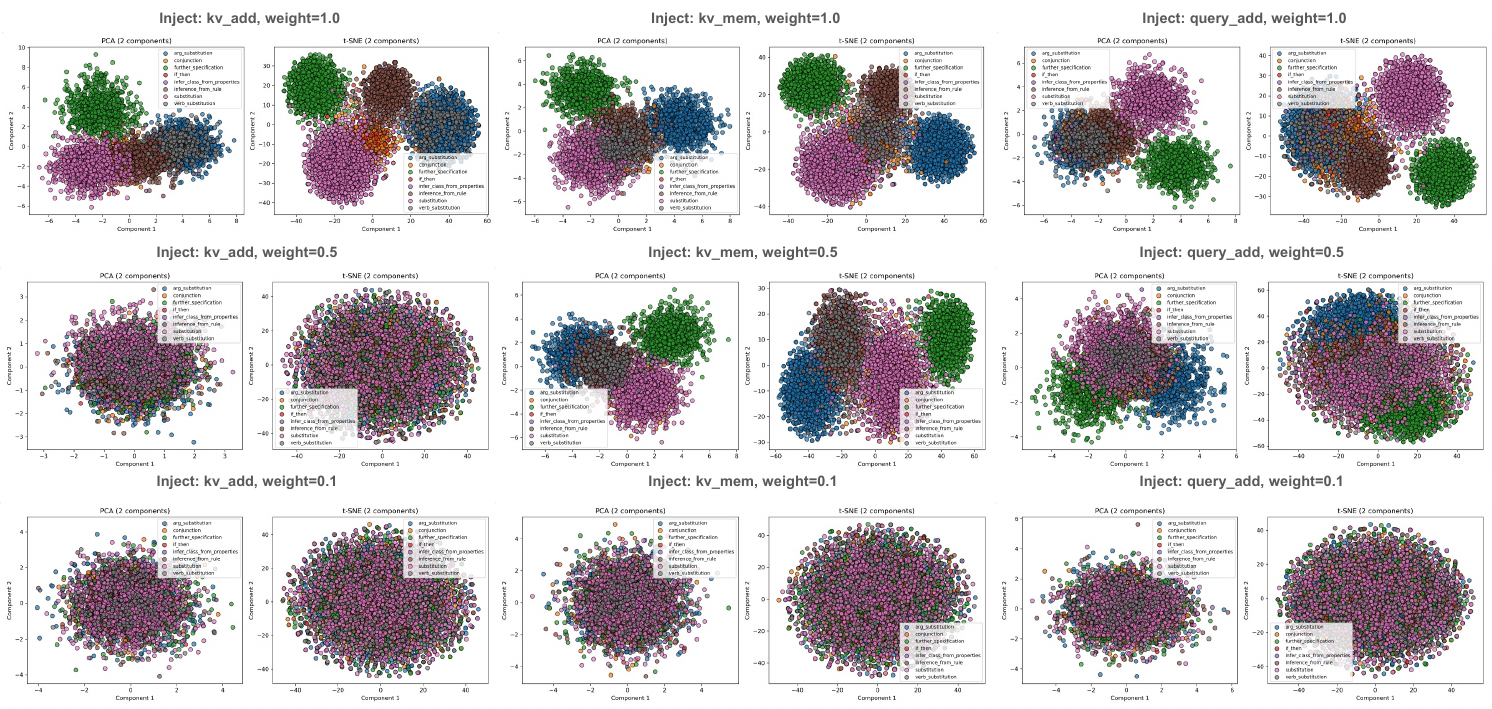}
    \caption{PCA and T-SNE visualisation for Explanatory Reasoning task.}
    \label{fig:explanation_pca}
\end{figure*}

\begin{figure*}[ht!]
    \centering
    \includegraphics[width=\linewidth]{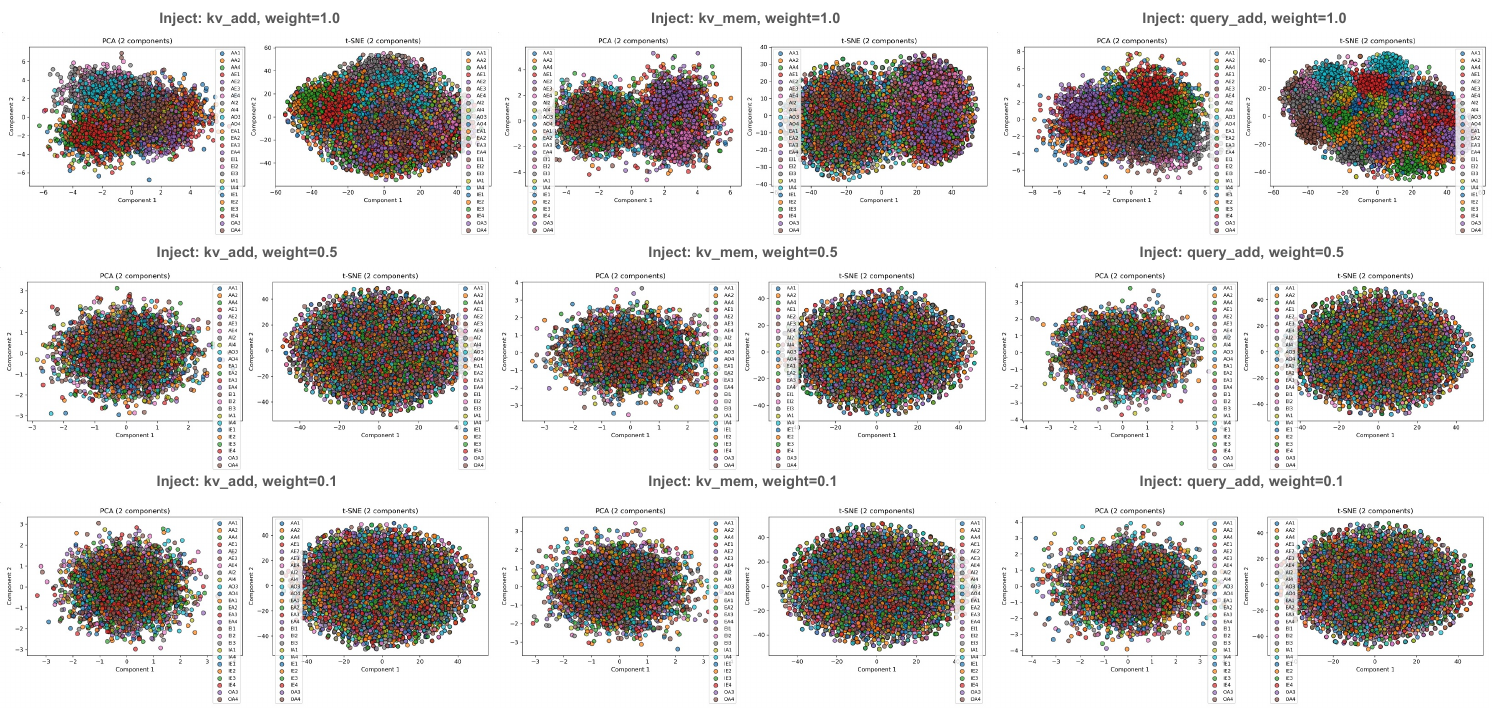}
    \caption{PCA and T-SNE visualisation for Syllogistic Reasoning task.}
    \label{fig:syllogistic_pca}
\end{figure*}

\paragraph{Feature space visualisation.} We provide the latent space visualisation across all reasoning tasks in Figure \ref{fig:math_pca}, \ref{fig:explanation_pca}, and \ref{fig:syllogistic_pca}.


\end{document}